\documentclass[runningheads]{llncs}

\usepackage{eccv}

\usepackage{eccvabbrv}
\usepackage{wrapfig}
\usepackage{graphicx}
\usepackage{graphicx}
\usepackage{booktabs}
\usepackage{multirow}
\usepackage[table]{xcolor}
\usepackage[accsupp]{axessibility}  % Improves PDF readability for those with disabilities.
\definecolor{pearDark}{HTML}{2980B9}

\usepackage[pagebackref,breaklinks,colorlinks,citecolor=eccvblue]{hyperref}
\usepackage{orcidlink}

\begin{document}
	
	% ---------------------------------------------------------------
	% TODO REVIEW: Replace with your title
	\title{OP-GRPO: Efficient Off-Policy GRPO for Flow-Matching Models} 
	
	% TODO REVIEW: If the paper title is too long for the running head, you can set
	% an abbreviated paper title here. If not, comment out.
	\titlerunning{OP-GRPO}
	
	% TODO FINAL: Replace with your author list. 
	% Include the authors' OCRID for the camera-ready version, if at all possible.
	\author{
		Liyu Zhang\inst{1,2} \and
		Kehan Li\inst{2}\textsuperscript{\textdagger} \and
		Tingrui Han\inst{2} \and
		Tao Zhou\inst{2} \and\\
		Yuxuan Sheng\inst{1} \and
		Shibo He\inst{1} \and
		Chao Li\inst{1}\textsuperscript{$\ast$}
	}
	
	% TODO FINAL: Replace with an abbreviated list of authors.
	\authorrunning{L. Zhang  et al.}
	% First names are abbreviated in the running head.
	% If there are more than two authors, 'et al.' is used.
	
	% TODO FINAL: Replace with your institution list.
	\institute{
		College of Control Science and Engineering, Zhejiang University\\
		\and Central Research Institute, Huawei\\
		\textsuperscript{$\ast$} Corresponding Author \quad
		\textsuperscript{\textdagger} Project Leader
	}
	\maketitle
	\begin{abstract}
		Post training via GRPO has demonstrated remarkable effectiveness in improving the generation quality of flow-matching models. However, GRPO suffers from inherently low sample efficiency due to its on-policy training paradigm. To address this limitation, we present OP-GRPO, the first \textbf{O}ff-\textbf{P}olicy \textbf{GRPO} framework tailored for flow-matching models. First, we actively select high-quality trajectories and adaptively incorporate them into a replay buffer for reuse in subsequent training iterations. Second, to mitigate the distribution shift introduced by off-policy samples, we propose a sequence-level importance sampling correction that preserves the integrity of GRPO's clipping mechanism while ensuring stable policy updates. Third, we theoretically and empirically show that late denoising steps yield ill-conditioned off-policy ratios, and mitigate this by truncating trajectories at late steps. Across image and video generation benchmarks, OP-GRPO achieves comparable or superior performance to Flow-GRPO with only 34.2\% of the training steps on average, yielding substantial gains in training efficiency while maintaining generation quality.
	\end{abstract}
	\section{Introduction}
	\label{sec:intro}
	
	Flow-matching models \cite{lipman2022flow,liu2022flow,ICLR2025_2460396f} have emerged as an effective paradigm for training diffusion models \cite{sohl2015deep,ho2020denoising,song2020score}, enabling continuous transport-based generative learning and significantly improving sampling efficiency and visual fidelity. Recent large-scale generative models built upon the flow-matching paradigm have significantly advanced modern visual generation. Representative models include image generation models such as Stable Diffusion 3.5 (SD3.5) \cite{esser2024scaling}, Flux \cite{labs2025flux1kontextflowmatching} and Qwen-Image \cite{wu2025qwenimagetechnicalreport}, as well as video generation models such as Wan \cite{wan2025}, Open-Sora 2.0 \cite{peng2025opensora20trainingcommerciallevel} and Kandinsky 5.0 \cite{arkhipkin2025kandinsky50familyfoundation}. 
	
	Building upon these generative backbones, downstream preference optimization methods leverage reinforcement learning (RL) signals to better align model outputs with human judgments \cite{black2023training,xu2023imagereward,wallace2024diffusion,liu2025flow,xue2025dancegrpo}. In particular, alignment approaches based on Group Relative Policy Optimization (GRPO) \cite{liu2025flow,xue2025dancegrpo} improve aesthetic quality, semantic consistency, and text rendering by optimizing group-normalized advantages over diffusion trajectories \cite{kirstain2023pick,ghosh2023geneval,wu2023human}.
	
	However, despite their alignment effectiveness, GRPO-based methods exhibit substantial inefficiency in large-scale flow-matching training \cite{li2025branchgrpo}. Existing approaches such as Flow-GRPO often require thousands of GPU hours to converge, substantially limiting scalability. This inefficiency stems primarily from two factors. First, GRPO follows a strictly on-policy paradigm: it repeatedly samples fresh trajectories under the current policy and discards them at the end of each policy iteration. As a result, even high-quality samples cannot be reused, leading to poor sample efficiency, while trajectory sampling itself constitutes a major portion of the overall computational cost. Second, when tasks are too difficult, the model rarely produces successful trajectories, causing rewards to degenerate toward all zeros. As a result, advantages collapse and gradient signals vanish, leaving the policy with little to learn from and stalling improvement.
	
	%Second, when sampled groups are either uniformly easy or uniformly difficult, the resulting rewards become degenerate (e.g., all ones or all zeros). In such cases, group-relative advantages collapse toward zero, producing weak gradient signals and slowing policy improvement.
	
	%However, existing GRPO-based methods, such as Flow-GRPO, suffer from severe efficiency issues and often require thousands of GPU hours to converge. This limitation primarily arises from two factors. First, GRPO is primarily on-policy, repeatedly collecting on-policy samples and updating the model based on these trajectories. Such samples are discarded after a single update, resulting in poor sample efficiency. The sampling procedure itself also accounts for a significant fraction of the total training cost. Second, when the samples are either excessively easy or overly challenging, the resulting samples tend to be homogeneous, producing degenerate rewards (all ones or all zeros). In such cases, the relative advantages computed within each group collapse to zero, leading to minimal gradient signals and consequently slow policy improvement.
	
	A natural solution is to introduce an off-policy training framework \cite{watkins1992q} that retains and reuses previously collected trajectories, thereby improving sample utilization and increasing training diversity. Off-policy learning has been widely studied in classical RL to enhance sample efficiency and stability \cite{mnih2013playing,haarnoja2018soft}. 
	
	%The most straightforward solution to this issue is to adopt an off-policy training framework, in which previously collected samples are retained and reused multiple times. Off-policy methods have been studied in traditional reinforcement learning (RL) to improve sample efficiency and increase diversity during training.
	
	% add description about the difference between diffusion model and LLM
	However, directly applying off-policy learning to flow-matching-based GRPO is non-trivial. Reusing trajectories generated by earlier policies induces distributional shift relative to the learned policy. This shift distorts the clipped objective in GRPO and may compromise training stability. Moreover, we observe that the severity of off-policiness varies significantly across denoising steps: as the process approaches low-noise regions, the conditional data distribution becomes highly concentrated, making importance weights increasingly ill-conditioned and exacerbating instability.

	%naively reusing off-policy samples for training diffusion models can introduce instability and may even degrade final performance. Off-policy samples introduce distributional shift with respect to both the sampling policy and the updated policy. This distributional shift leads to inaccuracies in the clipped objective of GRPO, thereby affecting training stability. Furthermore, we observe that the degree of off-policiness is strongly influenced by the denoising step: the farther the step is from pure noise, the more pronounced the off-policy effect, which similarly introduces instability during training.
	
	To address these challenges, we propose OP-GRPO: the first \textbf{O}ff-\textbf{P}olicy \textbf{GRPO} for Flow-Matching Models, an off-policy GRPO framework tailored for flow-matching generation models. OP-GRPO introduces sequence-level distribution correction to mitigate off-policy shift and employs denoising-step truncation to improve numerical stability during optimization. The main contributions of this paper are summarized as follows:
	
	%we incorporate an off-policy paradigm into flow-matching-based GRPO and introduce trajectory-based distribution correction along with truncated denoising-step measures to stabilize training. The main contributions of this paper are summarized as follows:
	
	(1) We maintain a replay buffer that  actively selects and retains high-quality off-policy trajectories. During rollout, a portion of on-policy samples are replaced with buffer samples to enable effective off-policy learning.
	
	(2) We introduce a sequence-level importance sampling correction that compensates for distributional discrepancies while preserving the clipping guarantees of GRPO, reducing clipped off-policy samples from over 40\% to 11.8\% and enabling stable updates.
	
	(3) We theoretically and empirically show that as noise level approaches zero, the transition distribution becomes nearly deterministic, causing ill-conditioned importance weights at late denoising steps. We therefore adopt a truncated strategy that excludes these late-stage off-policy samples to stabilize training.
	
	(4) Experiments are conducted on SD3.5-M for three tasks—compositional image generation, visual text rendering, and human preference alignment—and on Wan2.1-1.4B for visual text generation in video synthesis. OP-GRPO requires only approximately 34.2\% of the training steps needed by Flow-GRPO to reach its final performance, while achieving comparable or superior alignment metrics and generalization ability.

	\section{Related Work}
	\subsection{GRPO in Flow-matching Models}
	Recent studies on GRPO-based alignment for flow-matching models explore several algorithmic aspects. Flow-GRPO \cite{liuflow} and DanceGRPO \cite{xue2025dancegrpo} establish the paradigm of applying GRPO-style reinforcement learning to flow-matching generation through an ODE-to-SDE formulation. Building upon this stochasticization of flow dynamics, MixGRPO \cite{li2025mixgrpo} further generalizes sampling by adopting mixed ODE-SDE trajectory schedules to trade off transport fidelity and stochastic diversity during policy optimization. Subsequent work targets robustness of the policy update: BranchGRPO \cite{li2025branchgrpo} structures trajectory branching to separate multimodal denoising behavior, while GRPO‑Guard \cite{wang2025grpo} analyzes timestep-wise shifts in importance-ratio statistics and introduces ratio-normalization and gradient-reweighting corrections that restore effective clipping and prevent implicit over-optimization. Methods such as Fine-Grained GRPO \cite{zhou2025fine} and Neighbor GRPO \cite{he2025neighbor} refine credit assignment—via attribute-level decomposition or contrastive neighborhood objectives—to increase alignment granularity and preserve sample quality under preference supervision.
	
	Despite the above progress, existing methods are built on the on-policy nature of GRPO, where historical trajectories are discarded after training, leading to high computational cost and long training time. In contrast, our work extends GRPO in flow-matching models to an off-policy setting by enabling trajectory replay within policy optimization, thereby improving sample efficiency and reducing training cost.
	
	\subsection{Off-policy-based GRPO in LLM}
	Recent research has focused on extending GRPO beyond its original on-policy formulation to leverage historical experience and improve sample efficiency in large language model (LLM) post-training. RePO \cite{li2025repo} augments GRPO with a replay buffer that retrieves diverse off-policy samples for advantage estimation, increasing effective optimization steps without increasing computation and empirically improving reasoning performance over vanilla GRPO on multiple benchmarks. ReMix \cite{liang2025squeeze} generalizes this idea by enabling GRPO to leverage much larger volumes of off-policy data through a mix-policy proximal optimization framework that incorporates off-policy trajectories under KL-convex constraints, enabling stable reuse of historical data while preserving policy improvement guarantees. BAPO \cite{wan2026buffer} further proposes batch-level adaptation for selective reuse of high-value and difficult samples, providing theoretical insights into off-policy stability and empirical gains over standard GRPO. Liu et.al. \cite{liu-li-2025-rl-collapse} investigates the off-policy effects under verl \cite{sheng2024hybridflow} and Fully Sharded Data Parallel (FSDP) \cite{zhao2023pytorch} settings, and further explores several rollout correction methods.
	
	However, existing flow-matching-based GRPO work has not explored off-policy settings, resulting in low sample efficiency and slow convergence. Unlike autoregressive LLMs, flow-matching models parameterize a deterministic velocity field to transport noise to data. This mechanism fundamentally reshapes off-policy evaluation and distributional shift across timesteps. To this end, we present the first off-policy GRPO algorithm tailored to flow-matching models.

	\section{Preliminaries}
	\subsection{Flow-matching Models}
	Flow-matching models \cite{lipman2022flow,liu2022flow} learn a continuous transport from a base distribution $p_0$ (typically a standard normal distribution $\mathcal{N}(0,I)$) to the data distribution $p_1$ by parameterizing a time-dependent velocity field $v_\theta(x, t)$ that define the ordinary differential equation (ODE),
	\begin{equation}
		\frac{dx_t}{dt}=v_\theta(x, t),\ t \in [0, 1]
	\end{equation}
	
	This framework provides a simulation-free training paradigm that unifies diffusion-based generative models \cite{sohl2015deep,ho2020denoising,song2020score} and continuous normalizing flows \cite{chen2018neural}. During inference, samples are generated by numerically integrating the learned ODE, commonly implemented using an explicit Euler solver for efficient deterministic sampling.
	\begin{equation}
		\label{eqn:explicit Euler}
		x_{k-1} = x_k+\Delta t v_\theta(x_k, t_k)
	\end{equation}
	
	\subsection{On-policy GRPO}
	From a reinforcement learning perspective, this process can be interpreted as a trajectory, where each transition is governed by the rollout diffusion policy.
	
	The log-probability of the latent sequence can be decomposed as,
	\begin{equation}
		\text{log}\ p_{\text{off}}(\mathbf{z}_{0:T} \mid c) = \sum_{t=1}^{T}\text{log}\ p_{\text{off}}(\mathbf{z}_{t-1} \mid \mathbf{z}_t,\ c),
		\label{eqn:log_probs_decompose}
	\end{equation}
	which serves as a critical quantity for characterizing the underlying policy distribution and will be used to correct distributional mismatch in off-policy training via importance sampling.

	Reinforcement learning (RL) aims to learn a policy $\pi_\theta$ that maximizes the expected cumulative rewards with the following objective:
	\begin{equation}
		\text{max}_\theta \mathbb{E}_{(s_t, a_t)\sim \pi_\theta} \left[ \sum_{t=0}^T \big( R(s_t, a_t) - \beta D_{\text{KL}}(\pi_\theta (\cdot | s_t) || \pi_{\text{ref}}(\cdot | s_t))\big)\right],
	\end{equation}
	where $\pi_{\text{ref}}$ represents the reference policy and $D_{\text{KL}}$ measures the KL divergence and controls the update degree of learned policy $\pi_\theta$.
	
	Building on PPO algorithm \cite{schulman2017proximal}, GRPO estimates the advantage function with a group of samples. Formally, given a input prompt $c$, the flow model samples a group of images/videos $\{x_0^i\}_i^{G}$, with $G$ being the number of samples. The advantages of $i$-th image/video is calculated as follows:
	\begin{equation}
		\hat{A}^i_t = \frac{R(x_0^i, c) -\text{mean}(\{ R(x_0^j, c)_{j=1}^G \}) } {\text{std}(\{ R(x_0^j, c) \}_{j=1}^G)}.
	\end{equation}
	
	The objective of standard GRPO is,
	\begin{equation}
		\mathcal{J}(\theta)=\mathbb{E}_{c\sim\mathcal{C}, \{x_i\}_{i=1}^{G} \sim \pi_{\theta_{\text{old}}}(\cdot|c)}f(r, \hat{A}, \theta, \epsilon, \beta),
		\label{eqn:original_grpo_objective}
	\end{equation}
	where 
	\begin{equation}
		\nonumber
		\begin{aligned}
			f(r, \hat{A}, \theta, \epsilon, \beta) &= \frac{1}{G} \sum_{i=1}^G \frac{1}{T} \sum_{i=1}^{T-1} \Bigg( \text{min}\left( r_t^i \hat{A}^i_t, \text{clip}\left( r_t^i(\theta), 1-\epsilon, 1+\epsilon \right) \hat{A}^i_t \right)\Bigg),\\
			r^i_t(\theta) &= \frac{p_\theta(x_{t-1}^i | x_t^i, c)}{p_{\theta_{\text{old}}}(x_{t-1}^i | x_t^i, c)}.
		\end{aligned}
	\end{equation}
	
	\section{Method}
\begin{figure}[t]
		\begin{center}
			%\framebox[4.0in]{$\;$}
			\includegraphics[width=1.0\linewidth, trim=306 289 335 198, clip]{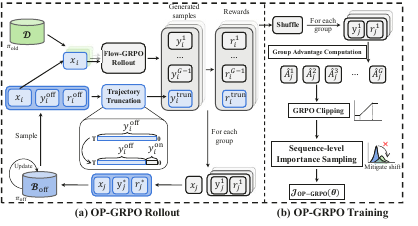} % 调整宽度并指定路径
		\end{center}
		\caption{Overall framework of OP-GRPO, including (a) OP-GRPO rollout and (b) OP-GRPO training. Blue regions represent samples from the replay buffer, green regions represent samples from the dataset.}
		\label{fig:overall_framework}
	\end{figure}

	In this section, we present OP-GRPO, an off-policy framework of GRPO designed to improve sample efficiency and accelerate training. We begin in \cref{sec:replay buffer} by introducing a high-quality replay buffer for storing off-policy trajectories with stepwise decay. In \cref{sec:training off-policy}, we formulate the off-policy training objective, with particular emphasis on sequence-level importance sampling for distribution correction. Finally, in \cref{sec: Truncated Denoising Steps}, we analyze the relationship between the degree of off-policy and the denoising steps, and propose truncated denoising to alleviate induced numerical instability in the optimization process. The overall framework of OP-GRPO is illustrated in \cref{fig:overall_framework} and refer to Appendix for the pseudocode.
	\subsection{High-quality Replay Buffer}
	\label{sec:replay buffer}
	\textbf{Replay Buffer Construction.} To preserve high-quality samples collected during the GRPO rollout phase, we maintain a replay buffer $\mathcal{B}_{\text{off}}$ to store these trajectories. During subsequent training iterations, a subset of stored samples is retrieved for reuse, thereby improving sample efficiency. Specifically, for each sampled group conditioned on $c$, we denote the set as $\{x_i\}_{i=1}^G$, where sample $x$ is formally defined as a triplet,
	\begin{equation}
		x = \big(c, \mathbf{z}_{0:T}, \text{log}\ p_{\text{off}}(\mathbf{z}_{0:T} \mid c), R\big),
		\label{eqn:training_sample}
	\end{equation}
	where $p_{\text{off}}$ denotes the distribution of diffusion policy, $\mathbf{z}_{0:T} = (\mathbf{z}_0, \mathbf{z}_1, \cdots, \mathbf{z}_T)$ denotes the complete latent trajectory generated by $\pi_{\text{off}}$, $\text{log}\ p_{\text{off}}(\mathbf{z}_{0:T} \mid c)$ denotes the log-probability of the generated trajectory and $R$ denotes the reward. 
	
	To ensure that $\mathcal{B}_{\text{off}}$ continuously retains the most informative and recent high-quality trajectories, we adopt a reward-based replacement strategy. Specifically, within each group, we select the highest-reward trajectory as the most informative sample for retention. We compare the newly selected trajectory with the lowest-reward entry currently stored in the buffer, and replace it if the new trajectory achieves a higher reward.
	
	Meanwhile, $\mathcal{B}_{\text{off}}$ enforces a uniqueness constraint to maintain the effectiveness and diversity of the stored trajectories. For each conditioning input $c$, at most one trajectory is retained. This design prevents the buffer from collapsing to a small set of trivial or repetitive samples and ensures that the stored data remains informative. When a newly selected trajectory corresponds to a conditioning input $c$ that already exists in the buffer, it is directly compared with the entry under the same condition, rather than with the lowest-reward entry.
	
	Furthermore, to ensure that samples in the buffer are effectively refreshed and to prevent excessively off-policy samples from being repeatedly introduced into training, the buffer $\mathcal{B}_{\text{off}}$ is designed to favor recently collected trajectories. This keeps the stored samples close to the current policy and and mitigates the risk of the training process becoming overly off-policy. To this end, we apply a stepwise decay to the rewards of trajectories in the buffer, so that older samples naturally become easier to replace, allowing newer, higher-quality trajectories to be incorporated more readily.
	
	%To enable the reuse of high-quality trajectories, we maintain a replay buffer $\mathcal{B}$ that stores trajectories with high rewards. Specifically, after collecting a new batch of on-policy samples, we evaluate them based on $R$ and determine whether they should be inserted. If the reward exceeds the minimum reward currently stored in $\mathcal{B}$, the trajectory is regarded as high-quality and replace the trajectory with the minimum reward. Otherwise, the sample is discarded. This reward-based replacement strategy ensures that the replay buffer continuously retains the most informative high-quality trajectories, facilitating effective off-policy training.
	
	\textbf{Rollout with buffer.} For the rollout stage of OP-GRPO, we adopt a hybrid RL paradigm, which follows the general philosophy of hybrid offline-to-online training \cite{song2022hybrid,nakamoto2023cal}. In each batch, prompts are composed of a majority of prompts drawn directly from the original dataset and a minority sampled from the buffer $\mathcal{B}_{\text{off}}$. This design enhances sample diversity while enabling effective reuse of high-value off-policy trajectories. Specifically, for prompts sampled directly from the dataset, $G$ trajectories are generated per prompt. For prompts sampled from the replay buffer, only $(G-1)$ new trajectories are generated, and the remaining trajectory is taken from the replay buffer. Formally, for a given input $c$, we construct the group as,
	\begin{equation}
		\mathcal{G}(c) = \Big\{ \mathbf{z}_{i}^{\text{on}}\sim p_{\text{old}}(\cdot|c) \Big\}_{i=1}^{G-1} \cup \Big\{\mathbf{z}^{\text{off}}\sim \mathcal{B}_{\text{off}} \Big\},
	\end{equation}
	where $G$ denotes the group size. These trajectories are then thoroughly shuffled before being used for subsequent training to prevent potential biases induced by a fixed sample ordering.
	
	% During training, a subsets of samples is directly drawn from the replay buffer $\mathcal{B}_{\text{off}}$. Since the replay buffer only stores individual high-quality samples rather than full groups, the remaining samples are generated online by the sampling policy, which corresponds to $p_{\text{old}}$. As a result, each group consists of two components: a single off-policy sample retrieved from the replay buffer and multiple on-policy samples generated by $p_{\text{old}}$.
	
	\subsection{Sequence-level Importance Sampling Correction}
	\label{sec:training off-policy}
	However, directly applying \cref{eqn:original_grpo_objective} to groups containing an off-policy sample is incorrect, as the training data is not sampled from the on-policy rollout policy $p_{\text{old}}$ (the policy at the beginning of the current update step), but rather from an off-policy policy $p_{\text{off}}$. This distributional mismatch causes biased gradient estimates, and if left uncorrected, the accumulated bias can destabilize training and cause the algorithm to diverge.
	
	A seemingly natural fix is to incorporate the distributional shift between $p_{\text{off}}$ and $p_{\text{old}}$ directly into the per-step importance sampling ratio in \cref{eqn:original_grpo_objective}. However, this approach is also inappropriate, as it fundamentally distorts the purpose of the clipping mechanism in GRPO. The clipping term is designed to bound the update magnitude of  $p_\theta$ \emph{relative to its starting point} $p_{\text{old}}$, thereby preventing excessively large policy updates and mitigate risk of instability or overfitting. However, once the distributional shift is naively absorbed into the per-step ratio, the $p_{\text{old}}$ terms cancel out, as shown in \cref{eqn:imporantce_ratio_cancel}:
	%On the other hand, it is also inappropriate to directly incorporate the distribution shift between $p_{\text{off}}$ and $p_{\text{old}}$ into the standard importance sampling ratio. Such a formulation actually measures the discrepancy between the current policy and the off-policy sampling distribution, while implicitly ignoring the role of $p_{\text{old}}$, as illustrated in ~\cref{eqn:imporantce_ratio_cancel},
	\begin{equation}
		r^i_t(\theta) = \frac{p_{\theta_{\text{old}}}(z_{t-1}^i | z_t^i, c)}{p_{\theta_{\text{off}}}(z_{t-1}^i | z_t^i, c)} \times \frac{p_\theta(z_{t-1}^i | z_t^i, c)}{p_{\theta_{\text{old}}}(z_{t-1}^i | z_t^i, c)}= \frac{p_\theta(z_{t-1}^i | z_t^i, c)}{p_{\theta_{\text{off}}}(z_{t-1}^i | z_t^i, c)}.
		\label{eqn:imporantce_ratio_cancel}
	\end{equation}
	
	As a consequence, the clipping term  measures the discrepancy between $p_\theta$ and $p_{\text{off}}$, a policy from potentially many iterations ago, rather than the intended reference policy $p_{\text{old}}$. Since these two policies can differ substantially, the importance sampling ratios become either very large or very small, causing a large fraction of samples to be clipped. Empirically, we find that this naive substitution results in over \textbf{40\%} of off-policy samples being clipped, which severely undermines the efficiency of off-policy learning and discards many otherwise useful off-policy samples.
	
	To address this issue, we propose a sequence-based correction scheme that simultaneously corrects for the distributional shift and preserves the intended behavior of the clipping mechanism. Specifically, etain the original per-step importance sampling ratio between $p_{\theta}$ and $p_{\text{old}}$
	, so that clipping continues to measure update magnitude relative to the correct reference policy. To compensate for the distributional shift introduced by the off-policy samples, we introduce a sequence-level correction term:
	\begin{equation}
		\mathcal{J}_{\text{OP-GRPO}}(\theta)=\mathbb{E}_{c\sim\mathcal{C}, \{x_i\}_{i=1}^{\mathcal{G}} }\ \frac{P_{\pi_{\text{old}}}(\tau)}{P_{\pi_{\text{off}}}(\tau)} \cdot  f(r, \hat{A}, \theta, \epsilon, \beta),
		\label{eqn:sequence_correlation_grpo_objective}
	\end{equation}
	where $f(r, \hat{A}, \theta, \epsilon, \beta)$ follows the same formulation as in \ref{eqn:original_grpo_objective} and $\frac{P_{\pi_{\text{old}}}(\tau)}{P_{\pi_{\text{off}}}(\tau)}$ is the sequence-level correction term defined below
	\begin{equation}
		\frac{P_{\pi_{\text{old}}}(\tau)}{P_{\pi_{\text{off}}}(\tau)} = \prod_{t=1}^{T}{\frac{p_{\theta_{\text{old}}}(z_{t-1}^i | z_t^i, c)}{p_{\theta_{\text{off}}}(z_{t-1}^i | z_t^i, c)}}.
	\end{equation}
	
	This formulation admits an intuitive interpretation from two complementary perspectives. If a sample is clipped, it indicates that the update of $p(\theta)$ relative to $p_{\text{old}}$ is excessively large. Such an aggressive update may lead to overfitting or other instability issues. Consequently, the sample is clipped, and the sequence-based correction term becomes inactive; the sample does not contribute to the update. In this case, the sequence-level correction term is inactive, and the sample does not contribute to the update. Conversely, when the update magnitude is moderate and the sample is not clipped, the correction term becomes active and compensates for the distributional shift introduced by the off-policy data, serving as an error-correction mechanism that ensures the unbiasedness of the policy update. Empirically, this formulation clips only 11.8\% of off-policy samples, significantly lower than the 40\% observed with the naive substitution, demonstrating that our approach effectively utilizes off-policy samples while maintaining training stability.
	
	\subsection{Trajectory Truncation for Numerical Stability}
	\label{sec: Truncated Denoising Steps}
	\begin{wrapfigure}{r}{0.45\linewidth}
		\vspace{-12pt}
		\centering
		\includegraphics[width=1.0\linewidth]{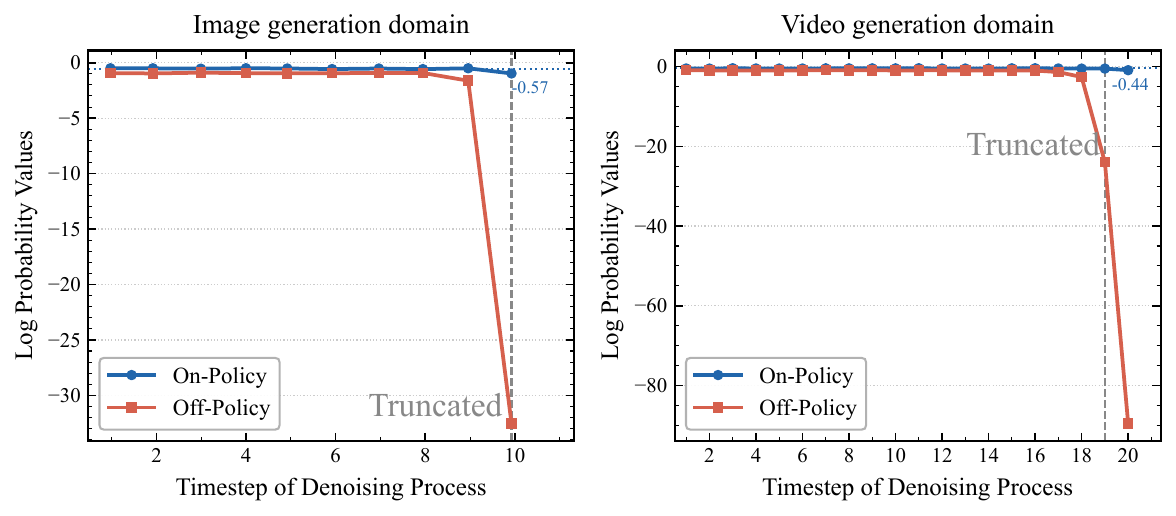}
		\label{fig:off-policy to denoising steps}
		\caption{Log-probability values of on-policy and off-policy samples across denoising steps, where the dashed line indicates the truncation starting step.}
		\vspace{-8pt}
	\end{wrapfigure}
	Our experiments show that, while incorporating off-policy data can effectively accelerate training, it also introduces noticeable instability. Unlike LLMs, where each token's log-probability is well-conditioned and roughly comparable in scale across positions \cite{ouyang2022training,liu2024deepseek}, flow-matching models exhibit a fundamentally different structure. The transition distribution $p_\theta(z_{t-1} | z_t, c)$ in flow-matching models becomes increasingly sharp as $\sigma \to 0$, causing log-probabilities at different denoising steps to exhibit substantially different numerical scales. This ill-conditioning can cause the importance weights to become disproportionately large or small in the later denoising steps, destabilizing policy updates.
	
	We verify this empirically by visualizing the log-probabilities of on-policy and off-policy samples across denoising steps, as shown in \cref{fig:off-policy to denoising steps}. The results show that the log-probability of off-policy samples remains relatively stable across most denoising steps, but undergoes a dramatic cliff-like drop in the final few steps. This confirms that the numerical ill-conditioning is concentrated in the low-noise regime, where the transition distribution becomes nearly deterministic.
	
	%To better understand this behavior, we visualize the log-probabilities of samples generated by the old policy and the off-policy across different denoising steps, as shown in \cref{fig:off-policy to denoising steps}. The results indicate that the log-probability of off-policy samples follows an approximately monotonic decreasing trend as denoising progresses, with a significantly sharper decline in the later steps. This observation suggests that, from a numerical perspective, the degree of off-policyness increases progressively with the denoising step.
	\begin{figure}[t]
		\begin{center}
			%\framebox[4.0in]{$\;$}
			\includegraphics[width=1.0\linewidth, trim=0 0 0 0, clip]{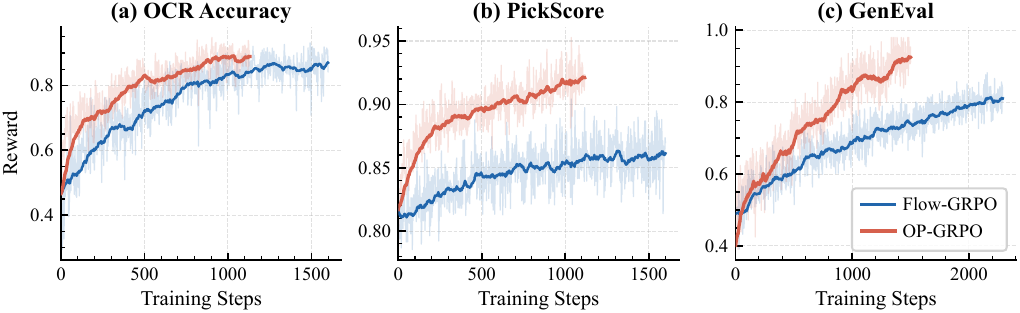} % 调整宽度并指定路径
		\end{center}
		\caption{\textbf{Training Curves of OP-GRPO and Flow GRPO.}}
		\label{fig:training_curves}
	\end{figure}
	
	We next analyze this phenomenon from a theoretical standpoint. The explicit form of the log-transition probability is:
	\begin{equation}
		\text{log}\ p_{\theta}(z_{t-1}^i | z_t^i, c) = \text{log}\ \mathcal{N}(z_{t-1}^i; \mu_t, \sigma_t) = -\frac{(z_{t-1}^i - \mu_t)^2}{2\sigma_t^2} - \text{log}\sigma_t - \frac{1}{2} \text{log}(2\pi),
		\label{eqn:log_prob_derivation}
	\end{equation}
	where $\mu_\sigma = z_t^i + v_\theta(x_t^i, \sigma) \Delta \sigma$ and $\sigma_t^2 = \sigma^2(\sigma_{prev} - \sigma)$. Refer to Appendix for detailed derivation.
	
	As $\sigma \to 0$, the transition variance $\sigma_t^2$ vanishes and the dynamics become nearly deterministic. In this regime, the quadratic term in \cref{eqn:log_prob_derivation} dominates and becomes highly sensitive to discretization and numerical errors, causing the importance weights to be numerically ill-conditioned and to disproportionately dominate the sequence-level correction term. This theoretical analysis is consistent with the empirical observation in \cref{fig:off-policy to denoising steps}. See Appendix for the detailed derivation.
	
	To address this issue, we adopt a simple yet effective trajectory truncation strategy: the off-policy trajectory is used only up to a truncation step $t_{\text{off}}$, after which the remaining steps are re-generated by the current policy:

	\begin{equation}
		\label{eqn:tau truncated step}
		\tau_{m} = \Big\{z^{\text{off}}_i\Big\}_{i=T}^{t_{\text{off}}} \cup \Big\{z_j\Big\}_{j=t_{\text{off}}}^{0},
	\end{equation}
	where $\tau_{m}$ denotes the mixed-policy trajectory, $z^{\text{off}}$ denotes the off-policy latents, $z_j$ denotes the latents re-generated by the current policy (as specified in \cref{eqn:explicit Euler}).
	
	This design directly eliminates the ill-conditioned log-probability terms in the low-noise regime, thereby stabilizing the importance weights and policy updates. Moreover, since the low-noise denoising steps contribute minimally to the perceptual quality of the generated image or video, this truncation introduces negligible degradation in generation quality, as confirmed by the training curves in \cref{fig:training_curves}.
	
	\section{Experiments}
	\begin{table*}[t]
		\centering
		\setlength{\tabcolsep}{4pt}
		\caption{Algorithm Performance on the \textbf{task metrics} including Compositional Image Generation, Visual Text Rendering, and Human Preference benchmarks. For the \textbf{unseen metrics} across image quality and preference score, ImgRwd denotes ImageReward and UniRwd denotes UnifiedReward. The best score is in \colorbox{pearDark!20}{blue}}
		\resizebox{\linewidth}{!}{
			\begin{tabular}{lccccccc}
				\toprule
				\multirow{2}{*}{\textbf{Model}} 
				& \multirow{2}{*}{\textbf{Step}}
				& \multirow{2}{*}{\textbf{Task Metric}} 
				& \multicolumn{2}{c}{\textbf{Image Quality}} 
				& \multicolumn{3}{c}{\textbf{Preference Score}}    \\
				\cmidrule(lr){4-5} \cmidrule(l){6-8} 
				& & 
				& \textbf{Aesthetic} 
				& \textbf{DeQA} 
				& \textbf{ImgRwd} 
				& \textbf{PickScore} 
				& \textbf{UniRwd} \\
				\midrule
				SD3.5-M & — & 0.63 / 0.59 / 21.72 & 5.31 & 4.07 & 0.92 & 22.18 & 3.29 \\
				\midrule
				\multicolumn{8}{c}{\textit{Compositional Image Generation}} \\
				\midrule
				Flow-GRPO & 1020 & 0.68 & 5.27 & 4.05 & 1.37 & 22.24 & 3.35 \\
				\rowcolor{gray!20} OP-GRPO   & 1020 & 0.88 & 5.22 & 4.02 & 1.21 & 22.35 & 3.49 \\
				Flow-GRPO & Best & 0.95 & 5.18 & 4.00 & 1.12 & 22.37 & 3.51 \\
				\rowcolor{gray!20} OP-GRPO   & Best & \colorbox{pearDark!20}{0.96} & 5.17 & 3.98 & 1.13 & 22.42 & 3.57 \\
				
				\midrule
				
				\multicolumn{8}{c}{\textit{Visual Text Rendering}} \\
				\midrule
				Flow-GRPO & 720 & 0.70 & 5.28 & 3.92 & 0.93 & 22.31 & 3.30 \\
				\rowcolor{gray!20} OP-GRPO   & 720 & 0.81 & 5.30 & 3.89 & 0.95 & 22.39 & 3.38 \\
				Flow-GRPO & Best & 0.92 & 5.32 & 3.83 & 0.98 & 22.54 & 3.46 \\
				\rowcolor{gray!20} OP-GRPO   & Best & \colorbox{pearDark!20}{0.93} & 5.32 & 3.85 & 0.98 & 22.56 & 3.45 \\
				\midrule
				
				\multicolumn{8}{c}{\textit{Human Preference Alignment}} \\
				\midrule
				Flow-GRPO & 720 & 22.19 & 5.44 & 4.13 & 1.09 & 22.19 & 3.43 \\
				\rowcolor{gray!20} OP-GRPO   & 720 & 23.27 & 5.92 & 4.17 & 1.22 & 23.27 & 3.59 \\
				Flow-GRPO & Best & 23.32 & 6.01 & 4.18 & 1.26 & 23.32 & 3.66 \\
				\rowcolor{gray!20} OP-GRPO   & Best & \colorbox{pearDark!20}{23.64} & 6.09 & 4.23 & 1.31 & 23.44 & 3.72 \\
				\bottomrule
			\end{tabular}
		}
		\vspace{-2mm}
		\label{tab:all_task}
	\end{table*}
	\subsection{Experimental Setup}
	To comprehensively evaluate of OP-GRPO framework, we conduct experiments across two generation models and three tasks.
	
	\textbf{Models.} We evaluate on \textbf{Stable-Diffusion-3.5-medium} (SD3.5-M)~\cite{esser2024scaling}, a state-of-the-art text-to-image model, and \textbf{Wan2.1-1.4B}~\cite{wan2025}, a recent text-to-video generation model, allowing us to assess the effectiveness of OP-GRPO across both image and video domains.
	
	\textbf{Tasks.} We consider three tasks, all following the experimental setup of Flow-GRPO: (1) \textbf{Compositional Image Generation}, which requires the model to accurately generate objects satisfying specified attributes such as count, color, and spatial relationships, evaluated with EvalGen~\cite{ghosh2023geneval}; (2) \textbf{Visual Text Generation}~\cite{chen2023textdiffuser}, which assesses the model's ability to render text explicitly specified in the prompt, evaluated via a rule-based recognition pipeline; and (3) \textbf{Human Preference Alignment}, which aims to align generation with human aesthetic preferences, using PickScore~\cite{kirstain2023pick} as the reward signal during training.
	
	%\textbf{Tasks.} We consider three tasks: (1) \textbf{Compositional Image Generation}, evaluated with EvalGen~\cite{ghosh2023geneval}; (2) \textbf{Visual Text Generation}~\cite{chen2023textdiffuser}, assessed via a rule-based text recognition pipeline; and (3) \textbf{Human Preference Alignment}, using PickScore~\cite{kirstain2023pick} as the reward model.
	
	\textbf{Generalization Evaluation.} To assess whether OP-GRPO generalizes beyond its training objectives, we report results on several unseen metrics. These metrics cover two dimensions: image quality, via Aesthetic Predictor~\cite{schuhmann2022laionaesthetics} and DeQA~\cite{you2025teaching}, and human preference, via ImageReward~\cite{xu2023imagereward}, PickScore~\cite{kirstain2023pick}, and UnifiedReward~\cite{wang2025unified}.

	%\textbf{Visual Text Generation.}\cite{chen2023textdiffuser} This task evaluates the model’s ability to generate high-quality visual content conditioned on textual prompts. We focus on core generation aspects, including visual fidelity, semantic alignment with the input text, and temporal consistency for video generation. Quantitative metrics and qualitative comparisons are conducted to assess improvements in visual quality and prompt adherence under the same inference settings.
	
	%\textbf{Human Preference Alignment.} This task measures how well the generated results align with human preferences. We perform both automatic reward-based evaluation and human studies, where annotators compare generated samples along dimensions such as aesthetic quality, text consistency, and overall preference. This setting directly reflects the effectiveness of preference-driven fine-tuning.
	
	%We evaluate on the following representative backbones,
	
	%We evaluate on the following representative backbones,
	
	%We evaluate on the following representative backbones,
	
	%We evaluate on the following representative backbones,

	\subsection{Training Efficiency of OP-GRPO}
	To demonstrate the training efficiency of OP-GRPO under the off-policy setting compared with Flow-GRPO, we visualize the training curves of both methods on the three aforementioned tasks, as shown in \cref{fig:training_curves}. OP-GRPO is trained for 1,150 steps, whereas Flow-GRPO runs for 1,600 steps. The results show that OP-GRPO requires only \textbf{34.2\%} of the training steps, on average, to reach the final performance achieved by Flow-GRPO. Moreover, OP-GRPO attains higher final performance on these tasks, with an average improvement of \textbf{1.17\%}. These results clearly demonstrate the superior training efficiency of OP-GRPO and highlight the advantages of incorporating off-policy learning.
	
	\begin{figure}[t]
		\begin{center}
			%\framebox[4.0in]{$\;$}
			\includegraphics[width=1.0\linewidth, trim=120 192 210 82, clip]{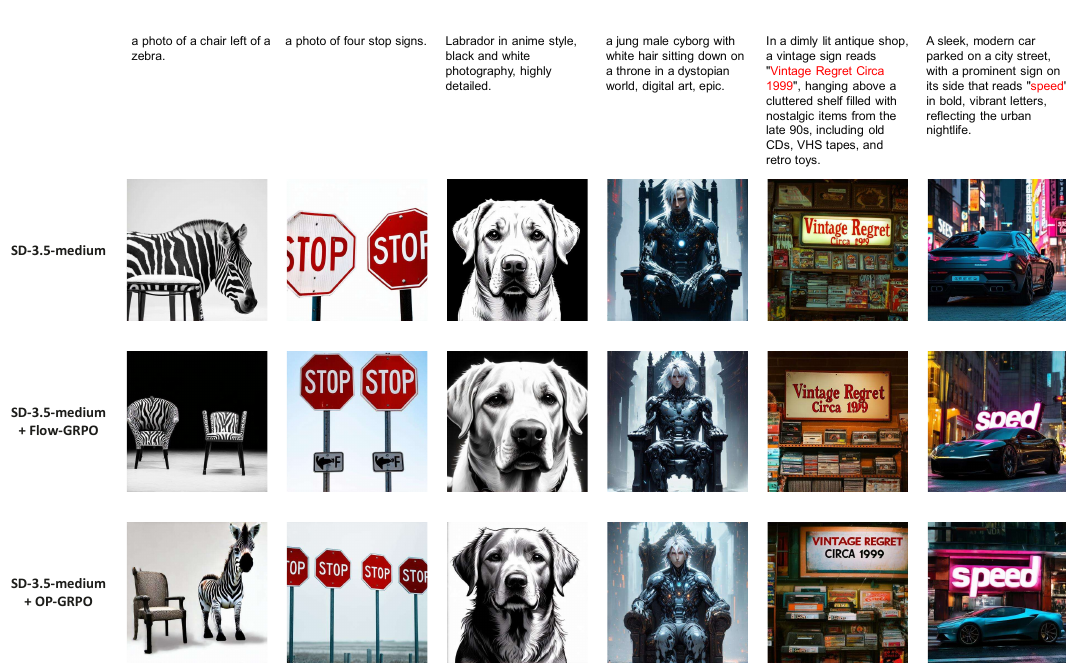} % 调整宽度并指定路径
		\end{center}
		\caption{Visual Results of OP-GRPO and Flow GRPO on three image generation tasks using SD3.5-M.}
		\label{fig:image_visualization}
	\end{figure}
	
	\subsection{OP-GRPO on Image Generation}
	\textbf{Quantitative Results.} We evaluate OP-GRPO on the three aforementioned tasks based on the {SD3.5-M} backbone, and summarize the results in \cref{tab:all_task}. We report both task-specific metrics (GenEval score for Compositional Image Generation, OCR accuracy for Visual Text Rendering, and PickScore score for Human Preference Alignment) and cross-task evaluation metrics. To comprehensively assess training dynamics, we present two groups of results: intermediate-step performance and the best-achieved performance during training. Please refer to Appendix for implementation details and hyperparameters.
	
	Overall, OP-GRPO consistently outperforms Flow-GRPO at intermediate training steps across all three tasks, not only on task-specific metrics but also on cross-task evaluation benchmarks. This consistent advantage indicates that OP-GRPO improves training efficiency without sacrificing generation quality or alignment properties. Regarding the best-achieved performance, OP-GRPO matches or slightly surpasses Flow-GRPO, suggesting that incorporating off-policy updates does not compromise the final optimization objective and can further enhance the upper-bound performance.
	
	\textbf{Visualization.} We further visualize intermediate-step generation results of Flow-GRPO and OP-GRPO on the three tasks, as shown in \cref{fig:image_visualization}. OP-GRPO demonstrates more precise control over object count and attributes in Compositional Image Generation, produces higher-quality and more legible text in Visual Text Rendering, and generates images with richer visual details and improved semantic coherence in Human Preference Alignment. These qualitative observations are consistent with the quantitative improvements reported in \cref{tab:all_task}.
	
	\textbf{Generalization Evaluation.} Beyond task-specific evaluation, we analyze the generalization capability of OP-GRPO from two perspectives. First, as shown in \cref{tab:all_task}, models trained on a particular task are evaluated using multiple cross-task metrics. OP-GRPO consistently achieves competitive or superior performance on these out-of-task benchmarks, indicating that the off-policy mechanism does not lead to overfitting to the task-specific reward, but instead promotes robust generation behaviors.
	
	Second, we further validate the scalability and extensibility of OP-GRPO on T2I-CompBench++ \cite{huang2023t2i,huang2025t2i}, a more comprehensive compositional text-to-image benchmark. The results shown in \cref{tab:t2i_comp_bench} demonstrate that OP-GRPO maintains clear advantages over Flow-GRPO under this more challenging and diverse evaluation protocol, confirming that the benefits of off-policy reuse extend beyond the original training tasks. Together, these findings suggest that OP-GRPO improves not only optimization efficiency but also generalization performance across varied generation benchmarks.
	\begin{table}[t]
		\centering
		\caption{{\bf T2I-CompBench++ Result.} We report results using the Best model trained on the GenEval. The best score is in \colorbox{pearDark!20}{blue}.}
		\resizebox{\linewidth}{!}{
			\begin{tabular}{lccccccc}
				\toprule
				\textbf{Model} & \textbf{Color} & \textbf{Shape} & \textbf{2D-Spatial} & \textbf{3D-Spatial} & \textbf{Numeracy} & \textbf{Non-Spatial} & \textbf{Average} \\ \midrule
				FLUX.1 Dev~\cite{flux2024}         & 0.738 & 0.574 & 0.289 & 0.391 & 0.621 & 0.315 & 0.488 \\
				SD3.5-M~\cite{esser2024scaling}    & 0.796 & 0.569 & 0.283 & 0.376 & 0.595 & 0.312 & 0.489 \\
				\midrule
				\multicolumn{8}{c}{\textit{\small Intermediate checkpoints}} \\
				\midrule
				SD3.5-M+Flow-GRPO            & 0.769 & 0.578 & 0.412 & 0.405 & 0.644 & 0.312 & 0.520 \\
				\textbf{SD3.5-M+OP-GRPO}     & 0.824 & 0.603 & 0.498 & 0.437 & 0.660 & 0.318 & 0.567 \\
				\midrule
				\multicolumn{8}{c}{\textit{\small Best checkpoints}} \\
				\midrule
				SD3.5-M+Flow-GRPO            & 0.838 & 0.613 & 0.545 & 0.447 & 0.675 & \colorbox{pearDark!20}{0.320} & 0.573 \\
				\textbf{SD3.5-M+OP-GRPO}     & \colorbox{pearDark!20}{0.845} & \colorbox{pearDark!20}{0.614} & \colorbox{pearDark!20}{0.550} & \colorbox{pearDark!20}{0.461} & \colorbox{pearDark!20}{0.682} & 0.319 & \colorbox{pearDark!20}{0.579} \\
				\bottomrule
			\end{tabular}
		}
		\vspace{-2mm}
		\label{tab:t2i_comp_bench}
	\end{table}
	
	\subsection{OP-GRPO on Video Generation}
\begin{figure}[t]
		\begin{center}
			%\framebox[4.0in]{$\;$}
			\includegraphics[width=1.0\linewidth, trim=15 219 82 110, clip]{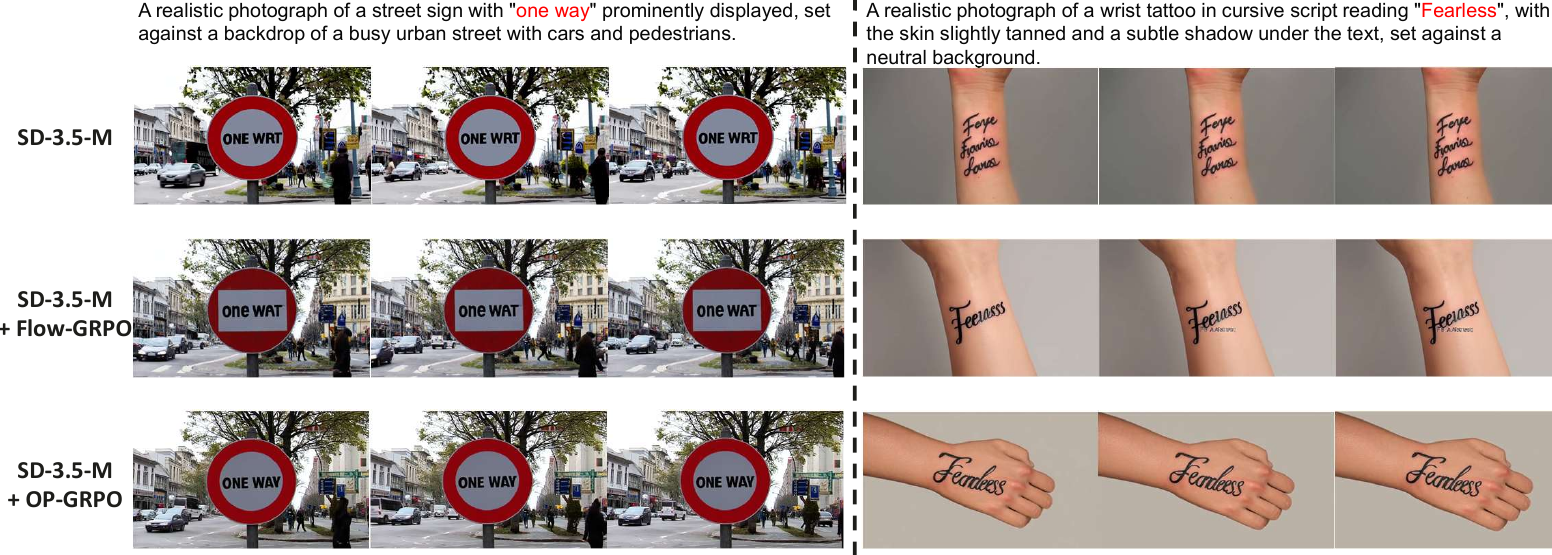} % 调整宽度并指定路径
		\end{center}
		\vspace{-5pt}
		\caption{Visual Results of Buffer-based GRPO and Flow GRPO on OCR task on video generation model Wan2.1-1.4B. Refer to Appendix for more results.}
		\vspace{-5pt}
		\label{fig:video_visualization}
	\end{figure}

	We further evaluate OP-GRPO on video generation to assess its scalability beyond image synthesis. Specifically, we conduct experiments on the Visual Text Generation task. \cref{fig:video_visualization} presents qualitative comparisons between OP-GRPO and Flow GRPO, while \cref{fig:training_curves} (d) illustrates the training convergence curves.
	
	The results demonstrate that OP-GRPO maintains strong performance in more challenging setting of video generation. Notably, OP-GRPO achieves the final performance of Flow GRPO using only 30.1\% of the training time on the video generation task, and its final generation quality substantially surpasses that of Flow GRPO. These findings highlight the effectiveness of OP-GRPO in addressing complex tasks and its superior efficiency in challenging scenarios.
	
	We attribute this gap to the limitation of the purely on-policy optimization adopted by Flow GRPO. For difficult tasks, such a training paradigm is often hard to bootstrap: even when high-quality samples are occasionally discovered, they are immediately discarded, preventing the model from sufficiently reinforcing beneficial generation patterns, which leads to slow and unstable improvement. In contrast, OP-GRPO retains these informative samples and repeatedly leverages them through off-policy updates, enabling more effective credit assignment and sustained exploitation of high-quality trajectories. This design results in superior performance in complex video generation scenarios.
	
	\subsection{Ablation Study}
\begin{figure}[t]
		\begin{center}
			%\framebox[4.0in]{$\;$}
			\includegraphics[width=1.0\linewidth, trim=0 0 0 0, clip]{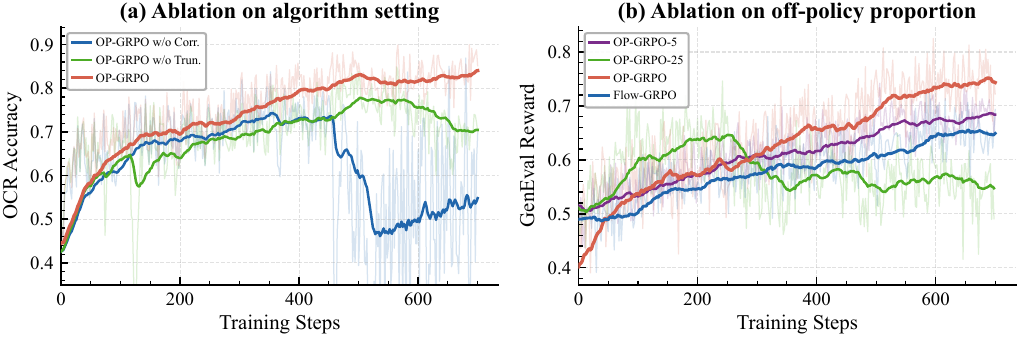} % 调整宽度并指定路径
		\end{center}
		\vspace{-5pt}
		\caption{Ablation study of OP-GRPO.}
		\vspace{-5pt}
		\label{fig:ablation study}
	\end{figure}

	\textbf{Algorithm settings.} We investigate the impact of key algorithmic components on performance. Specifically, we evaluate several variants of our method, including (1) a variant without the sequence correction term (denoted as \textit{OP-GRPO w/o corr}), and (2) a variant without truncating the denoising steps (denoted as \textit{OP-GRPO w/o trun}). The results are shown in ~\cref{fig:ablation study} (a). Without the sequence correction term, the algorithm becomes highly unstable, as the distribution shift accumulates over training and eventually leads to divergence, clearly demonstrating the necessity of this component. Without truncation, similarly suffers from instability, since excessively off-policy samples introduce severe distribution mismatch that poses significant challenges to policy updates. In contrast, OP-GRPO remains stable and unbiased, ensuring efficient and robust optimization.
	
	\textbf{Impact of Off-policy Sample Proportion.} We further investigate how the proportion of off-policy samples affects algorithm performance. As shown in \cref{fig:ablation study} (b), we evaluate two variants with off-policy proportions of 5\%, and 25\%, denoted as OP-GRPO-5 and OP-GRPO-25, respectively. The results show that incorporating off-policy samples consistently accelerates convergence regardless of the proportion used. However, when the proportion is too low, the benefit is marginal: OP-GRPO-5 achieves only a 23.1\% improvement in convergence speed. On the other hand, a higher proportion introduces more high-quality off-policy samples and thus leads to faster convergence, but at the cost of increased instability or even divergence, as larger distribution shifts make policy updates more susceptible to interference. Therefore, selecting an appropriate off-policy sample proportion according to the characteristics of the task is crucial for balancing convergence speed and training stability.

	\section{Conclusion}
	In this work, we investigate the efficiency and stability bottlenecks of GRPO-based RL training for flow-matching diffusion models. We identify that the on-policy training paradigm and degenerate reward signals significantly hinder learning efficiency, while naive off-policy reuse introduces distributional shift and training instability. To address these issues, we propose OP-GRPO, an off-policy GRPO framework that integrates replay-based sampling, sequence-level distribution correction, and truncated denoising. Experiments on image and video generation benchmarks demonstrate that OP-GRPO substantially accelerates convergence while preserving generation quality competitive with fully on-policy baselines. One limitation is that our current evaluation focuses on image and video generation tasks; extending OP-GRPO to other domains, such as audio or 3D generation, remains an open direction for future work.

	%OP-GRPO provides a practical and principled step toward more efficient preference alignment for large-scale diffusion models. On limitation 

	\newpage
	% ---- Bibliography ----
	%
	% BibTeX users should specify bibliography style 'splncs04'.
	% References will then be sorted and formatted in the correct style.
	%
	\bibliographystyle{splncs04}
	\bibliography{main}
\end{document}